\documentclass[runningheads]{llncs}
\usepackage[T1]{fontenc}
\def\doi#1{\href{https://doi.org/\detokenize{#1}}{\url{https://doi.org/\detokenize{#1}}}}
\usepackage{graphicx}
\usepackage{color}
\usepackage{graphicx}
\usepackage{amsmath}
\usepackage{newtxmath}
\usepackage{amsfonts}
\usepackage{multirow}
\usepackage{tabularx}
\usepackage{ulem}
\usepackage{bm}
\usepackage[dvipsnames]{xcolor}
\usepackage{threeparttable}
\usepackage{amsmath}
\usepackage[pagebackref=true,breaklinks=true,colorlinks,bookmarks=false]{hyperref}
\usepackage{cite}
\usepackage{ragged2e}
\usepackage{booktabs}
\usepackage{color}

\begin{document}
%

\title{Disease-informed Adaptation of Vision-Language Models}

\author{Jiajin Zhang \inst{1}
\and Ge Wang \inst{1}
\and Mannudeep K. Kalra \inst{2}
\and Pingkun Yan \inst{1}
}


\authorrunning{J. Zhang et al.}
\institute{Department of Biomedical Engineering and Center for Biotechnology and Interdisciplinary Studies, Rensselaer Polytechnic Institute, Troy, NY, USA\\
\and Department of Radiology, Massachusetts General Hospital,
Harvard Medical School, Boston, MA, USA\\ 
}

%
%
%
\maketitle              
\begin{abstract}

In medical image analysis, the expertise scarcity and the high cost of data annotation limits the development of large artificial intelligence models. This paper investigates the potential of transfer learning with pre-trained vision-language models (VLMs) in this domain. Currently, VLMs still struggle to transfer to the \textit{underrepresented} diseases with minimal presence and \textit{new} diseases entirely absent from the pretraining dataset. We argue that effective adaptation of VLMs hinges on the nuanced representation learning of disease concepts. By capitalizing on the joint visual-linguistic capabilities of VLMs, we introduce disease-informed contextual prompting in a novel disease prototype learning framework. This approach enables VLMs to grasp the concepts of new disease effectively and efficiently, even with limited data. Extensive experiments across multiple image modalities showcase notable enhancements in performance compared to existing techniques. The code will be made publicly available on our github page: \url{https://github.com/RPIDIAL/Disease-informed-VLM-Adaptation}.

\end{abstract}
\section{Introduction}

In medical image analysis, the scarcity of required expertise and the high cost of data annotation hinder the development of deep learning models~\cite{annotation_01,annotation_02,zhang2023neural,zhang2023spectral,zhang2023toward,luan2024spectrum,luan2023high,zhang2023revisiting}. Thus, transfer learning of pre-trained models emerged as a practical solution. Recent advancements have seen pre-trained vision language models (VLMs), such as the CLIP-based networks~\cite{clip,clip_nat,plip,ms_cxr}, achieved impressive domain adaptation capabilities. The integration of natural language processing allows VLMs to learn better visual representations with an appropriate alignment with textual concepts, thereby facilitating model generalizability. However, the recent works~\cite{vlm_lack_trans, vlm_lack_trans_2} showed that the efficacy of VLMs diminishes when the relevant category are only sparsely represented in the training data.
This paper attempts to address two major challenges in adapting VLMs in computer-assisted diagnosis with medical images, including \textit{underrepresented diseases} with minimal presence and \textit{new diseases} entirely absent from the pretraining dataset.

In a clinical environment, it is often feasible to obtain a limited number of data related to an emerging or different disease for model adaptation. For example, during the early stage of the COVID-19 pandemic, a selection of image samples were rapidly shared to spread awareness and understanding of the disease~\cite{rsna_covid}.
Consequently, employing data-efficient methods for adapting pretrained large models emerges as a viable path for medical image analysis.
Several methods have been proposed for transfer learning with VLMs. Adapter-based approaches like linear probing~\cite{clip}, CLIP-Adapter~\cite{clip_adapter}, and Tip-Adapter~\cite{tip_adapter} adopt an classifier to tailor pre-trained visual encoders on new tasks, by only tweaking the final layers. 
Alternatively, prompting-based methods like CoOp~\cite{coop} and CoCoOp~\cite{cocoop} focus on optimizing learnable prompts without actual text inputs, prioritizing performance over the acquisition of meaningful concepts. The recent method MaPLe~\cite{maple} utilizes dual-modality prompts to adapt CLIP to new domains, yet it demands extensive data for finetuning and may not suitable for diseases with limited data. In this paper, we propose a new framework of disease-informed adaptation to address the above limitations.

Our research contends that effective representation learning of disease concepts is central to successfully adapting vision language models (VLMs). Our method innovates in the following two fronts.
%
\textbf{1)} \textit{Disease-informed Contextual Prompting} (DiCoP) utilizes the clinical knowledge to craft prompts for representing the concepts of the target diseases. The prompts highlight the disease characteristics with descriptive attributes, such as texture, shape, and location. To overcome the problem that crafted prompts lack instance specifics, we further propose to enhance the textual prompts with image context features.
\textbf{2)} \textit{Disease Prototype Learning} (DPL) addresses the lack of structural regulation in the latent space of CLIP-based VLMs~\cite{cyclip}, which is critical for recognizing the target diseases. The DPL framework fine-tunes the image encoder to actively learn the prototypes of diseases, and regularizes the geometric structure of the learned representations for the downstream visual recognition tasks.

\begin{figure}[t]
    \centering
    \includegraphics[width=\columnwidth, clip=true, trim=0 2 0 2]{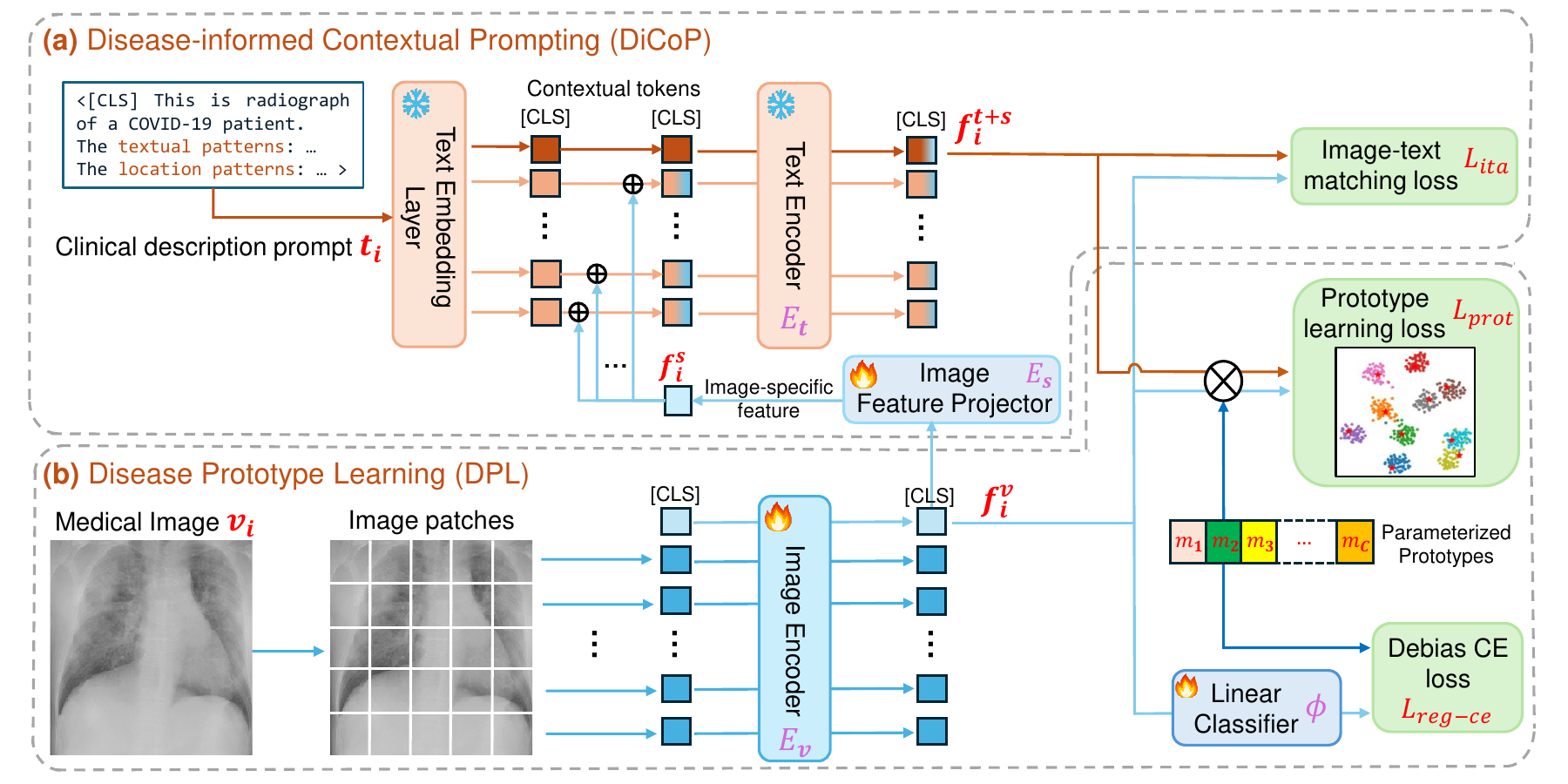}
    \caption{The framework overview. 
    \textbf{(a)} DiCoP produces prompts informed by specific diseases. \textbf{(b)} DPL enables learning disease representations with limited data.
}
    \label{fig:method_overview}
\end{figure}

To demonstrate the effectiveness and efficiency of our proposed approach, we carried out extensive experiments across various imaging modalities and diagnostic tasks. Our evaluation specifically targets two illustrative use cases. In the first scenario, focused on identifying cancerous pathological images, the target cancer type is \textit{underrepresented} due to much less data compared to other cancer types in the pretraining dataset. The second scenario mirrors a clinical situation, where a \textit{newly identified} disease was not present in the pretraining dataset with only a few samples available for model adaptation. An example of this type is the COVID-19 pandemic's onset. Our method can help improve the preparedness and response when we deal with future health crisis.

\section{Method}

Our method utilizes medical VLMs, pretrained in the CLIP style, to diagnose underrepresented or new diseases. The core idea of our method is the effective representation learning of disease concepts. This involves two steps. First, \textit{Disease-informed Contextual Prompting} (DiCoP) is designed to bridge the concepts of new diseases with established clinical knowledge and instance-specific features.
Then, \textit{Disease Prototype Learning} (DPL) is utilized to learn disease prototypes by imposing geometric regularization on the learned representations. 


As shown in Fig.~\ref{fig:method_overview}, a CLIP model comprises a text encoder $\boldsymbol{E}_t$ and an image encoder $\boldsymbol{E}_v$. 
We denote a labeled downstream dataset of target diseases as $\{(\boldsymbol{v}_i, \boldsymbol{y}_i)\}_{i=1}^N$. The images $\boldsymbol{v}_i$ are grouped into $C$ categories, each represented by a distinct disease or condition. 
Our proposed DiCoP method utilizes clinical knowledge to generate text prompts $\boldsymbol{t}_i$ for each image. The image and text encoders, $\boldsymbol{E}_v$ and $\boldsymbol{E}_t$, process each $(\boldsymbol{v}_i, \boldsymbol{t}_i)$ pair, using the \texttt{[CLS]} token to derive global visual and linguistic representations $\boldsymbol{f}^{t}_{i}, \boldsymbol{f}^{v}_{i} \in \mathbb{R}^{h \times 1}$, respectively. We \textbf{freeze} the pretrained text encoder to prevent it from overfitting to the small set of hand-crafted prompts.




\subsection{Disease-informed Contextual Prompting (DiCoP)}

Drawing on insights from the prior research on natural images analysis~\cite{prompt_diversity,vlm_lack_trans,vlm_lack_trans_2,cocoop}, we underscore the value of incorporating both \textit{clinical knowledge} and \textit{image-specific features} into contextual prompts to enhance the transferability of VLMs. The key lies in linking concepts of underrepresented/new diseases with the existing clinical knowledge. For instance, radiological descriptions of texture attributes can aid in illustrating pneumonia. Through this approach, the concept of pneumonia is connected to the pre-existing radiology corpus via linguistic semantics by the text encoder $\boldsymbol{E}_t$.
To this end, we propose crafting contextual prompts containing notable attributes of target categories. In particular, we develop descriptive contextual prompts using the template
\begin{equation}
\label{eq:prompt_template}
\texttt{Prompt}_{k} |_{k=1}^C = 
\texttt{Des}_{k}(\texttt{texture}) 
\oplus 
\texttt{Des}_{k}(\texttt{location}) 
\oplus 
\texttt{Des}_{k}(\texttt{shape}),
\end{equation}
%
where $\oplus$ symbolizes the concatenation of descriptions $\texttt{Des}_{k}(\cdot)$ for the $k_{th}$ disease category, covering each attribute (texture, location, shape). Each description $\texttt{Des}_{k}(\cdot)$ is manually designed with clinical knowledge. The predefined prompts are fixed for all samples of the same category, \textit{i.e.}, $\boldsymbol{t}_i=\texttt{Prompt}_{k}$ if $i\in \boldsymbol{S}_k$. 

Given the visual diversity of images within the same category, we follow CoCoOp~\cite{cocoop} to enrich text-only prompts in Eq. \ref{eq:prompt_template} with image-specific features. Fusing general clinical knowledge and specific image variability leads to accurate contextual prompts. As depicted in Fig.~\ref{fig:method_overview}, an image feature projector $\boldsymbol{E}_s$ is introduced to extract image-specific representations $\boldsymbol{f}^{s}_{i} = \boldsymbol{E}_s(\boldsymbol{f}^{v}_{i})$, where $\boldsymbol{f}^{v}_{i}$ is the visual \texttt{[CLS]} token encoded by $\boldsymbol{E}_v$. The dimensionality of image-specific representation $\boldsymbol{f}^{s}_{i}$ matches the contextual token embeddings. $\boldsymbol{f}^{s}_{i}$ is then added to each contextual token embeddings, except for the \texttt{[CLS]} token, to enhance the prompts with image-specific information. 
We denote the representation of text-only prompts as $\{\boldsymbol{f}^{t}_{k}\}_{k=1}^C$, and the representations augmented with image-specific features as $\{\boldsymbol{f}^{t+s}_{i}\}_{i=1}^N$. An image-text alignment loss $L_{ita}$ is defined as follows to encourage the model to maximize the cosine similarity between the matched images and prompts, while reducing the similarity between unmatched pairs. $\tau_{1}$ is a temperature hyperparameter.
\begin{small}
\begin{equation}
\label{eq:it_matching_loss}
L_{ita} = \frac{1}{2N} \sum_{i=1}^{N} \left[ 
\frac{\exp(\boldsymbol{f}^{v}_{i} \cdot \boldsymbol{f}^{t+s}_{i}/\tau_{1})}{\log \left( \sum_{j=1}^{N} \exp(\boldsymbol{f}^{v}_{j} \cdot \boldsymbol{f}^{t+s}_{i}/\tau_{1}) \right)} 
+ 
\frac{\exp(\boldsymbol{f}^{v}_{i} \cdot \boldsymbol{f}^{t+s}_{i}/\tau_{1})}{\log \left( \sum_{j=1}^{N} \exp(\boldsymbol{f}^{v}_{i} \cdot \boldsymbol{f}^{t+s}_{j}/\tau_{1}) \right)} \right]
\end{equation}
\end{small}

\subsection{Disease Prototype Learning (DPL)}

By employing DiCoP generated prompts, which combines clinical knowledge and image-specific feature, alongside a limited number of new image samples, we fine-tune medical VLMs to learn representations of underrepresented or new diseases. However, the original CLIP-based VLMs do not impose any geometric constraints on the representations within each modality, which is a limitation~\cite{cyclip} for transferring VLMs to the downstream disease diagnosis tasks. To address the challenge, we propose a framework that enhances the explicit representation learning for the disease of each category.

More specifically, we employ trainable parameters $\{\boldsymbol{m}_k\}^C_{i=1}$ to represent the $C$ distinct disease prototypes. Each prototype $\boldsymbol{m}_k$ is initialized by $\boldsymbol{f}^{t}_{k}$, which is the representation of the text-only $\texttt{Prompt}_k$ without image-specific features.
Following Eq.~\ref{eq:prototype_loss}, we fine-tune the model and the prototypes simultaneously by minimizing the cosine similarity for samples grouped under the same category and their respective prototype $\boldsymbol{m}_k$. Concurrently, we maximize the separability between different disease prototypes.
\begin{small}
\begin{equation}
\label{eq:prototype_loss}
L_{prot} = \sum_{k=1}^c \frac{1}{|2 S_k|} \sum_{\boldsymbol{i} \in S_k} \left[\exp\left(\frac{\boldsymbol{f}^{v}_{i} \cdot \boldsymbol{m}_k}{\tau_{2}}\right) + \exp\left(\frac{\boldsymbol{f}^{t+s}_{i} \cdot \boldsymbol{m}_k}{\tau_{2}}\right)\right]
- 
\lambda_1 \sum_{k \neq j} \exp \left(\frac{\boldsymbol{m}_k \cdot \boldsymbol{m}_j}{\tau_{2}}\right)
\end{equation}
\end{small}
where $\tau_{2}$ is another temperature hyperparameter and $\lambda_1$ is a weighting factor.

As in Fig.~\ref{fig:method_overview}, the linear classifier $\phi$ classifies an image with prediction probabilities $\boldsymbol{p}_{i} = \phi(\boldsymbol{f}^{v}_{i})$.
To avoid overfitting the learned disease prototypes $\boldsymbol{m}_k$ to the scarce samples, we further employ the disease-informed prompts, specified in Eq.~\ref{eq:prompt_template}, as concept anchors to regularize the $\boldsymbol{m}_k$. Subsequently, we incorporate a regularized cross-entropy loss as
\begin{small}
\begin{equation}
\label{eq:debias_ce_loss}
L_{reg-ce} = 
-\frac{1}{N}\sum_{i=1}^{N} \log(\boldsymbol{p}_{i}) \cdot \boldsymbol{y}_{i}
+
\frac{\lambda_2}{C}\sum_{k=1}^{C} ||\boldsymbol{m}_k - \boldsymbol{f}^{t}_k ||_2\ ,
\end{equation}
\end{small}
where $\lambda_2$ is a weighting factor. The regularization term avoids excessive model adaptation to the limited samples by preserving a minimal $\ell_2$-norm distance between the disease prototypes $\boldsymbol{m}_k$ and the representations $\boldsymbol{f}^{t}_k$ of the disease-informed textual prompts. This helps align the model's learning closer to clinical knowledge rather than being narrowly tailored to the limited data. $\boldsymbol{E}_v$, $\boldsymbol{E}_s$, $\phi$ and $\{\boldsymbol{m}_k\}^C_{i=1}$ are optimized simultaneously by minimizing the loss $L_{total} = L_{ita} + L_{prot} + L_{reg-ce}$. Only the vision branch $\boldsymbol{E}_v$ and $\phi$ are used for the inference.

\section{Experiments and Results}

We evaluated the proposed method on two datasets, PanNuke and COVID-x. \textbf{PanNuke}~\cite{pannuke} contains 7,558 pathological images for 19 organs. The task is a binary classification (malignant or benign) based on the tissue appearances. 
When applied to downstream tasks, PLIP showed suboptimal zero-shot performance (F1<0.5) on certain organs, including the kidney, colon, and pancreas, because of their \textit{under-representation} in the pretraining dataset~\cite{plip}. We followed PLIP~\cite{plip} to resize images to $224\times224$ pixels, and divided the data into training/validation/test sets with the ratio of $7:1:2$.
\textbf{COVID-x}(v6)~\cite{covidx} contains chest x-ray images collected from multiple countries. The task is a 3-way classification (COVID-19, non-COVID pneumonia, or normal). 
BioViL~\cite{ms_cxr}, pretrained on the MIMIC dataset~\cite{mimic}, was developed prior to the COVID pandemic. COVID-19 is thus a \textit{new} disease for the model. Since we do not have access to the test set, we split the original training set of 29,634 images into train and validation subsets with a ratio of $7:1$, and used the original validation set of 400 images for test. We followed BioViL to resize all images to $512\times 512$ pixels. 

\subsection{Implementation Details}

\noindent\textbf{Network Architectures:} PLIP uses ViT-B/16~\cite{vit} as its image encoder, while BioViL utilizes ResNet-50~\cite{resnet}.\footnote{The text encoders are frozen. Since it is computationally expensive to directly tune the vision encoders, we adopted LoRA~\cite{lora} (rank $r=4$ and scaling $\alpha=4$) to fine-tune PLIP on PanNuke. The BioViL was finetuned on COVID-x using layer-wise learning rate decay~\cite{layer_rate_decay} (decay factor $\beta=0.9$).} The image feature projector $\boldsymbol{E}_s$ employs a two-layer bottleneck architecture (Linear-ReLU-Linear), compressing the input dimension by a factor of $16$. The classifier consists of a simple linear layer, mapping the image features to $K$ classes (class number $K \leq$  prototype number $C$).

\noindent\textbf{Hyperparameters and training:} All models and prototypes are trained for $40$ epochs with a batch size of $64$, using a learning rate of $0.0005$ and the AdamW optimizer. We set the temperature parameters $\tau_1=\tau_2=0.07$ and weighting factors $\lambda_1 = 
\lambda_2 = 0.1$ in all the experiments.

\noindent\textbf{Disease description and prompt preparation:} GPT-4~\cite{gpt4} was employed to draft $\texttt{Prompt}_{k}$ with the template outlined in Eq.~\ref{eq:prompt_template}, followed by subsequent manual revisions under the supervision of \textit{a clinical professional}. The prompt examples are given in Supplementary Sec. 1.
For the PanNuke dataset, a total of 38 $\texttt{Prompt}$ templates were created to represent benign and malignant tissues across 19 organs, corresponding to 38 unique learnable disease prototypes. 
For the COVID-x dataset, 3 $\texttt{Prompt}$ templates and 3 trainable disease prototypes were developed to depict COVID-19, non-COVID pneumonia, and normal cases respectively.
\begin{table}[t]
\centering
\caption{10-run average AUC and F1 scores of models on PanNuke with $5\%$ training data. The best performance is in \textbf{bold} and the second best is \underline{underlined}.}
\scalebox{0.75}{
\begin{threeparttable}
    \begin{tabular}{p{95pt}<{\centering} p{45pt}<{\centering} p{25pt}<{\centering} p{45pt}<{\centering} p{25pt}<{\centering} p{45pt}<{\centering} p{25pt}<{\centering} p{55pt}<{\centering} p{25pt}<{\centering}}
        \toprule
        \multirow{2}{*}{\textbf{Method}} & \multicolumn{2}{c}{\textbf{Kidney}} & \multicolumn{2}{c}{\textbf{Colon}} 
        & \multicolumn{2}{c}{\textbf{Pancreatic}} & \multicolumn{2}{c}{\textbf{All Organs}} \\
        \cmidrule(lr){2-3} \cmidrule(lr){4-5} \cmidrule(lr){6-7} \cmidrule(lr){8-9}
        & F1 & AUC & F1 & AUC & F1 & AUC & Weighted F1 & AUC \\
        %
        \midrule
        Linear Probing & 0.560$^*$ & 0.660$^*$ & 0.748$^*$ & 0.794$^*$ & 0.810$^*$ & 0.754$^*$ & 0.740$^*$ & 0.779$^*$\\
        CLIP-Adapter & 0.587$^*$ & 0.703$^*$ & 0.774$^*$ & 0.853$^*$ & 0.853$^*$ & 0.812$^*$ & 0.786$^*$ & 0.825$^*$\\
        Tip-Adapter & 0.610$^*$ & 0.722$^*$ & 0.788$^*$ & 0.859$^*$ & 0.877$^*$ & 0.830$^*$ & 0.802$^*$ & 0.843$^*$\\
        \midrule
        CoOp & 0.617$^*$ & 0.730$^*$ & 0.810$^*$ & 0.863$^*$ & 0.874$^*$ & 0.828$^*$ & 0.804$^*$ & 0.847$^*$ \\
        CoCoOp & \underline{0.647}$^*$ & \underline{0.753}$^*$ & 0.826$^*$ & 0.882$^*$ & \underline{0.903}$^*$ & \underline{0.861} & \underline{0.838}$^*$ & \underline{0.880}$^*$ \\
        MaPLe & 0.635$^*$ & 0.745$^*$ & \underline{0.836}$^*$ & \underline{0.891} & 0.891$^*$ & 0.853$^*$ & 0.827$^*$ & 0.869$^*$ \\
        \midrule
        DiCoP + DPL (ours) & \textbf{0.692} & \textbf{0.786} & \textbf{0.860} & \textbf{0.897} & \textbf{0.920} & \textbf{0.875} & \textbf{0.865} & \textbf{0.896} \\
        \bottomrule
    \end{tabular}
    
    \begin{tablenotes}\footnotesize
    \item[$*$] $p<0.05$ in the one-tailed paired samples Student's \textit{t}-test with 10-run results using our method.
    \end{tablenotes}
\end{threeparttable} 
}
\label{tab:pannuke}
\end{table}

\subsection{Method Effectiveness}
We demonstrate the effectiveness of our method by comparing it with adapter-based methods (linear probing~\cite{clip}, CLIP-Adapter~\cite{clip_adapter} and Tip-Adapter~\cite{tip_adapter}), prompting-based approaches (CoOp~\cite{coop}, CoCoOp~\cite{cocoop} and MaPLe~\cite{maple}). 

\noindent\textbf{PanNuke.} When applied to downstream tasks, PLIP shows suboptimal zero-shot performance ($F1$<0.5) on certain organs, including the kidney, colon, and pancreas, which were due to their \textit{under representation} in the pretraining dataset as mentioned by the authors~\cite{plip}. We fine-tuned the PLIP by sampling $5\%$ of the training data in each run and calculated the 10-run mean metric values on the test set to eliminate the randomness.
The results are presented in Table~\ref{tab:pannuke}. 
The adapter-based methods generally performed the worst among all methods. This indicates that simply tweaking the last layer may not effectively leverage the VLMs’ capabilities in the downstream task.
%
Our approach significantly ($p<0.05$ with Student’s t-test) outperformed other prompting-based methods in almost all the cases. This underscores DiCoP's advantage over optimizing prompts without clinical knowledge, and demonstrates DPL's benefit in tailoring VLMs to less represented organs, thereby avoiding overfitting on scarce data.

 \begin{figure}[t]
    \centering
    \includegraphics[width=\columnwidth, clip=true, trim=0 5 0 2]{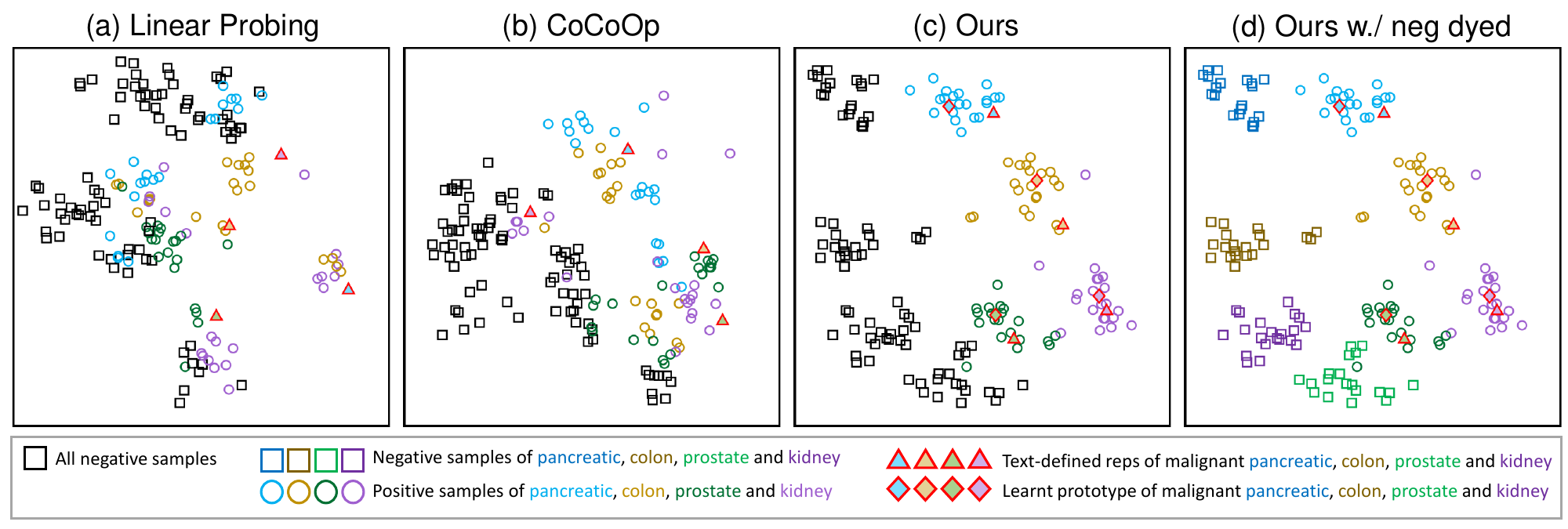}
    \caption{t-SNE visualization on the latent space on PanNuke. \textbf{(a)} Linear probing, \textbf{(b)} CoCoOp, \textbf{(c)} our method, and \textbf{(d)} ours with negative samples of each organ dyed distinctly.
}
    \label{fig:results_representation}
\end{figure}

Fig.~\ref{fig:results_representation} visualizes the geometric structure of the learned representations of different models via t-SNE~\cite{tsne}. 
Compared to \textbf{(a)} linear probing and \textbf{(b)} CoCoOp, our method in \textbf{(c)} produced encouraging results in superior separation between positive and negative samples. That explains the excellent performance in Table.~\ref{tab:pannuke}.
In addition, our method presents compact organ-specific clusters, which is attributed to DPL. The clustered representations can be also extended to negative samples by coloring each organ differently as shown in \textbf{(d)}. In contrast to our method, CoCoOp does not properly align the positive samples ($\circ$) of each organ with the representations of their corresponding text-only prompt ($\triangle$). This misalignment indicates CoCoOp's weakness in lacking the clinical knowledge from the textual input and geometric regularizations in the latent space.

\begin{table}[h]
\centering
\caption{10-run average results of COVID finding and all findings on COVID-x with 1\% training data. The best performance is in \textbf{bold} and the second best is \underline{underlined}.}
\scalebox{0.72}{
\begin{threeparttable}
    \begin{tabular}{p{95pt}<{\centering} p{80pt}<{\centering} p{80pt}<{\centering} p{80pt}<{\centering} p{80pt}<{\centering}}
        \toprule
        \multirow{2}{*}{\textbf{Method}} & \multicolumn{2}{c}{\textbf{COVID-19 finding}} & \multicolumn{2}{c}{\textbf{All findings}}\\
        \cmidrule(lr){2-3} \cmidrule(lr){4-5}
        & Precision & Recall 
        & Acc & Weighted F1 \\
        \midrule
        Linear Probing & 0.668$^*$ & 0.601$^*$ & 0.623$^*$ & 0.621$^*$\\
        CLIP-Adapter & 0.675$^*$ & 0.645$^*$ & 0.638$^*$ & 0.640$^*$\\
        Tip-Adapter & 0.682$^*$ & 0.667$^*$ & 0.640$^*$ & 0.641$^*$\\
        \midrule
        CoOp & \underline{0.665}$^*$ & 0.628$^*$ & 0.638$^*$ & 0.636$^*$\\
        CoCoOp & 0.603$^*$ & 0.703$^*$ & 0.662$^*$ & 0.658$^*$\\
        MaPLe\textdagger & - & - & - & -\\
        \midrule
        DiCoP + DPL (ours) & \textbf{0.702} & \textbf{0.770} & \textbf{0.741} & \textbf{0.744}\\
        \bottomrule
    \end{tabular}
    
    \begin{tablenotes}\footnotesize
    \item[$*$] $p<0.05$ in the one-tailed paired Student's \textit{t}-test with 10-run results of our method. 
    \item[\textdagger] MaPLe cannot directly be applied on the CNN-based VLM. 
    \end{tablenotes}
\end{threeparttable} 
}
\label{tab:covid}
\end{table}

\noindent\textbf{COVID-x.} BioViL~\cite{ms_cxr}, pretrained on the MIMIC dataset~\cite{mimic}, was developed prior to the COVID pandemic. This makes COVID-19 a \textit{new} disease for the model. Similarly, we did 10-run training with $1\%$ random sampling of the training data. Compared to other methods, Table~\ref{tab:covid} shows the effectiveness of our method with significantly higher average precision and recall on detecting COVID-19, as well as significantly better accuracy and weighted-F1 scores across all classes.
The results underscore the benefits of explaining diseases with clinical language by DiCoP and disease concept learning via DPL.

\vspace*{-0.5\baselineskip}
\subsection{Data Efficiency}

\begin{figure}[t!]
    \centering
    \includegraphics[width=\columnwidth, clip=true, trim=0 2 0 2]{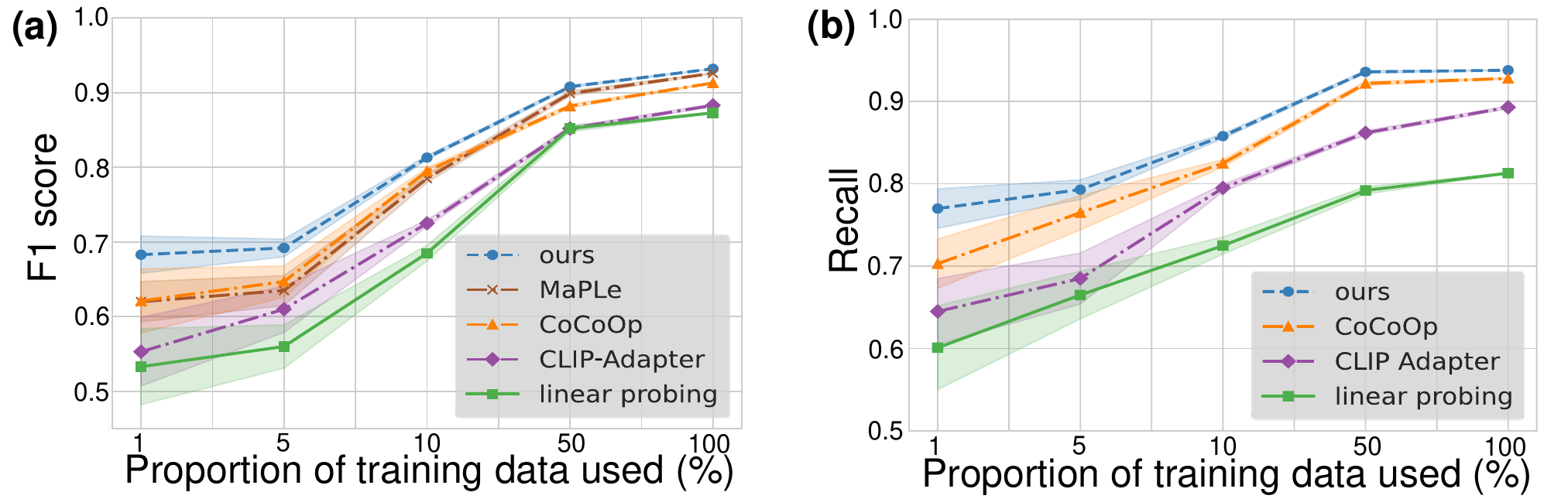}
    \caption{Data efficiency on \textbf{(a)} kidney cancer in PanNuke and \textbf{(b)} COVID in COVID-x.
}
    \label{fig:results_data_efficiency}
\end{figure}

Considering the data scarcity of \textit{underrepresented} or \textit{newly emerged} diseases, we examine data efficiency by testing the model performance with different proportions of training data. Fig.~\ref{fig:results_data_efficiency}~\textbf{(a)} and \textbf{(b)} display our method's superior 10-run average F1 score of the kidney tissue classification in PanNuke and recall ratio of the COVID class in COVID-x, respectively.
Notably, our approach surpasses other baselines, particularly with limited training data.

\begin{figure}[t!]
    \centering
    \includegraphics[width=\columnwidth, clip=true, trim=0 2 0 2]{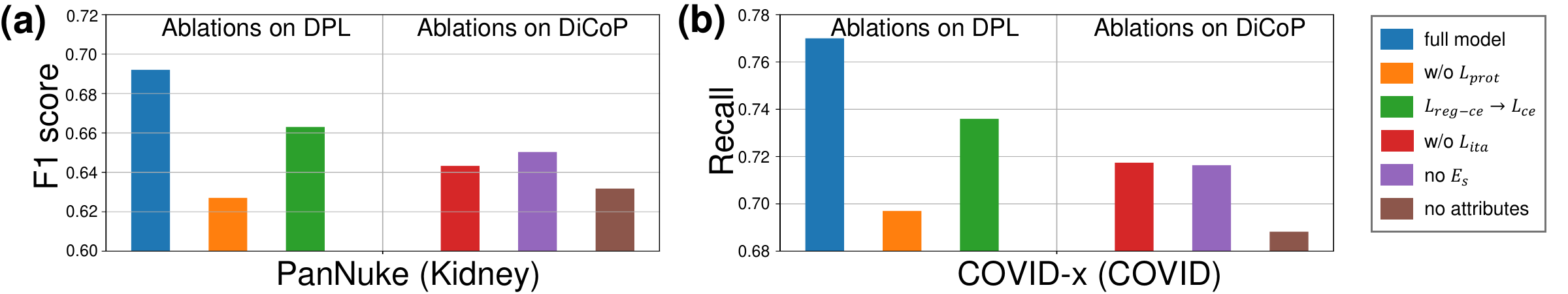}
    \caption{Ablation studies on \textbf{(a)} kidney in PanNuke and \textbf{(b)} COVID in COVID-x.
}
    \label{fig:results_ablation_study}
\end{figure}

\vspace*{-0.5\baselineskip}
\subsection{Ablation Study}
To evaluate the efficacy of the each component in our method, we conducted an ablation study on the kidney tissue classification in PanNuke (5\% training data) and COVID detection in COVID-x (1\% training data)
As shown in Fig.~\ref{fig:results_ablation_study}, the study includes: \textbf{1)} excluding each loss (w/o $L_{prot}$, w/o $L_{ita}$); \textbf{2)} replacing $L_{reg-ce}$ with normal CE loss ($L_{reg-ce}\rightarrow L_{ce}$); \textbf{3)} no image-specific features (no $E_{s}$); \textbf{4)} removing the descriptive attributes from Eq.~\ref{eq:prompt_template} (no attributes). Each study was done in a leave-one-out fashion. Fig.~\ref{fig:results_ablation_study} demonstrates a decline in model performance relative to the full model upon removing or altering each component, indicating the efficacy of all components.

\section{Conclusion}

This paper introduces a novel framework for adapting pretrained VLMs to \textit{underrepresented} or \textit{new} diseases when data is limited. Empirical analyses validate the effectiveness and efficiency of our method across different VLMs and tasks. Our approach facilitates the disease-informed computer-aided diagnosis, potentially broadening its application to other medical image analysis tasks.

\section*{Acknowledgments}

This research was partially supported by the National Science Foundation (NSF) under the CAREER award OAC 2046708, the National Institutes of Health (NIH) under award R21EB028001.



%
%
%
%
\newpage
\bibliographystyle{splncs04}
\bibliography{refs}

\end{document}